# GPT-3 Models are Poor Few-Shot Learners in the Biomedical Domain

**Milad Moradi, Kathrin Blagec, Florian Haberl, Matthias Samwald**
Institute for Artificial Intelligence
Medical University of Vienna, Austria
{milad.moradivastegani, kathrin.blagec, matthias.samwald}@meduniwien.ac.at,
florian.haberl@live.at

## Abstract

Deep neural language models have set new breakthroughs in many tasks of Natural Language Processing (NLP). Recent work has shown that deep transformer language models (pretrained on large amounts of texts) can achieve high levels of task-specific few-shot performance comparable to state-of-the-art models. However, the ability of these large language models in few-shot transfer learning has not yet been explored in the biomedical domain. We investigated the performance of two powerful transformer language models, i.e. GPT-3 and BioBERT, in few-shot settings on various biomedical NLP tasks. The experimental results showed that, to a great extent, both the models underperform a language model fine-tuned on the full training data. Although GPT-3 had already achieved near state-of-the-art results in few-shot knowledge transfer on open-domain NLP tasks, it could not perform as effectively as BioBERT, which is orders of magnitude smaller than GPT-3. Regarding that BioBERT was already pretrained on large biomedical text corpora, our study suggests that language models may largely benefit from in-domain pretraining in task-specific few-shot learning. However, in-domain pretraining seems not to be sufficient; novel pretraining and few-shot learning strategies are required in the biomedical NLP domain. Codes to run the experiments are available at https://github.com/mmoradi-iut/BioGPT-3.

## 1 Introduction

Recent developments in transformer-based deep neural language models such as BERT (Devlin et al., 2018), XLNet (Yang et al., 2019), and GPT-3 (Brown et al., 2020) have led to new breakthroughs in many Natural Language Processing (NLP) tasks. These high-performance NLP models take the advantage of task-agnostic pretraining on massive text corpora to encode lexical, syntactic, and semantic regularities of the language. Task-specific fine-tuning is then used to master task-specific language capabilities such as named entity recognition, coreference resolution, relation extraction, semantic similarity estimation, etc. Recent research has demonstrated that scaling up an autoregressive language model, i.e. GPT-3 with 175 billion parameters, can lead to achieving near state-of-the-art performance in task-specific few-shot learning, without the need for fine-tuning on the whole training data (Brown et al., 2020). However, it is not yet known whether GPT-3 can perform well on domain-specific texts, e.g. biomedical text, in few-shot settings.

Domain-specific language models such as BioBERT (Lee et al., 2019), SciBERT (Beltagy et al., 2019), and Clinical XLNet (Huang et al., 2020) were pretrained on very large biomedical and clinical text corpora; they achieved state-of-the-art results after task-specific fine-tuning on (bio)medical datasets. However, the ability of these large language models in few-shot learning has not yet been explored.

In this paper, we investigate the few-shot transfer learning abilities of GPT-3 in biomedical NLP (BioNLP) tasks. This helps to understand whether GPT-3 can effectively perform in BioNLP use cases where only a limited number of training samples are available. We also evaluate the performance of a pretrained biomedical language model, i.e. BioBERT, in few-shot settings to figure out whether large language models are proficient few-shot learners in the biomedical domain.

We conducted extensive experiments on various BioNLP tasks, i.e. textual inference, relation classification, semantic similarity estimation, question answering, and text classification, in few-shot settings. The results revealed that, in few-shot settings, the large language models could not achieve performance scores remotely competitive with those of a language model fine-tuned on the whole training data. The experimental results suggest that, in the BioNLP domain, there is still much room for development of multi-task language models that can effectively transfer knowledge to new tasks where a small amount of training data is available.

## 2 BioNLP tasks

We evaluated GPT-3 and BioBERT on five different biomedical and clinical NLP tasks in few-shot settings.

**MedNLI** (Romanov and Shivade, 2018) is a textual inference dataset containing more than 14K pairs of sentences, such that every pair belongs to a class `entailment`, `contradiction`, or `neutral`, depending on the semantic relation between the two sentences.

**BioText** (Rosario and Hearst, 2004) is a relation classification dataset that contains more than 3.5K text snippets from Medline abstracts. Every sample is classified into one of eight classes based on the relationship between disease and treatment entities mentioned in the text.

**MedSTS** (Wang et al., 2020) is a semantic similarity dataset consisting of more than 1K pairs of sentences from the Mayo Clinics clinical text database. A score between 0 and 5 is assigned to each pair.

**PubMed-RCT** (Dernoncourt and Lee, 2017) contains 200K PubMed abstracts annotated with sentence classifications. Each sentence is classified into one of the classes `Background`, `Objective`, `Method`, `Result`, or `Conclusion` based on the role of sentence in the abstract.

**PubMed-QA** (Jin et al., 2019) is a question answering dataset containing more than 1K research questions extracted from PubMed abstracts, along with short and long answers.

The evaluation measure for MedNLI and PubMed-QA was **Accuracy**. BioText and PubMed-RCT were evaluated using **Micro-F1**. **Pearson Correlation** was used to measure the performance on MedSTS.

All the experiments were performed on a computer with an Intel Core i5-9600K CPU at 3.70GHz, 32 GB of RAM, and a GeForce RTX 2080 Ti graphic card (GPU) with 11 GB dedicated memory.

## 3 Experimental results

We tested the GPT-3-derived models `Curie` and `Davinci` available through the OpenAI API[1]. For brevity reasons, we only report the results obtained by `Curie`, however, we got similar results when the more complex `Davinci` model was used. In the first round of experiments, we used the OpenAI API general-purpose built-in function `Completion`, which can be used for various tasks. It receives a prompt that contains a textual description of the task and few training samples, such that every sample has a context and a desired completion to instruct the model on how the output should look like. In the second round of experiments, we used two other built-in functions designed for specific tasks. We used the `Classification` function to run the model on **MedNLI**, **BioText**, **MedSTS**, and **PubMed-CT** tasks. This function receives a query and a set of labeled samples, searches through the samples to find the most relevant ones, and predicts the most likely label for the query. To perform the **PubMed-QA** task, we used the `Answer` function which answers an input question based on the provided context and a set of training samples.

Table 1 presents the results of evaluating GPT-3 on the five BioNLP tasks in few-shot settings. The scores achieved by fine-tuned BioBERT-`Large` are also presented in this table, in order to compare few-shot learning performance against fine-tuning on the full training set. The results are presented for different numbers of few-shot learning samples randomly drawn from the training set. The samples were stratified in order to have an even class distribution over few-shot learning examples. Since there was a maximum context window size of 2048 tokens when we used `Completion` function, it was not possible to use more than 50 training samples as context in most tests. However,

---

[1] https://beta.openai.com/

| GPT-3 built-in function | #training samples | BioNLP tasks | | | | |
|---|---|---|---|---|---|---|
| | | MedNLI | BioText | MedSTS | PubMed-RCT | PubMed-QA |
| Completion | 5 | 0.36 | 0.31 | 0.19 | 0.28 | 0.51 |
| | 10 | 0.43 | 0.48 | 0.27 | 0.44 | 0.59 |
| | 20 | 0.38 | 0.62 | 0.42 | 0.39 | 0.56 |
| | 50 | 0.41 | 0.57 | 0.38 | 0.47 | 0.56 |
| | 100 | * | * | * | * | * |
| Classification | 5 | ** | ** | ** | ** | - |
| | 10 | ** | ** | ** | ** | - |
| | 20 | 0.30 | ** | ** | ** | - |
| | 50 | 0.32 | ** | ** | ** | - |
| | 100 | 0.40 | ** | 0.11 | 0.43 | - |
| | 200 | 0.37 | 0.53 | 0.26 | 0.41 | - |
| Answer | 5 | - | - | - | - | ** |
| | 10 | - | - | - | - | ** |
| | 20 | - | - | - | - | 0.42 |
| | 50 | - | - | - | - | 0.54 |
| | 100 | - | - | - | - | 0.59 |
| | 200 | - | - | - | - | 0.51 |
| Fine-tuned BioBERT-Large | Full training set | 0.85 | 0.91 | 0.84 | 0.92 | 0.81 |

Table 1: Performance of GPT-3 on the BioNLP tasks in few-shot settings using three different built-in functions. *The length of prompt exceeds the model's context window that equals 2048 tokens. **The model could not find enough relevant training samples for several test samples, therefore, no scores were produced because of a model's error.

there was no such limitation when we used Classification and Answer functions.

Two hyperparameters *max_examples* and *max_rerank* were used by functions Classification and Answer, respectively, to control the number of relevant samples and documents when predicting the most probable labels and answers. For brevity reasons, only the results for *max_examples*=5 and *max_rerank*=5 are reported in Table 1. There was no significant change in the scores when these hyperparameters took other values in the range [2, 10]. However, we report evaluation scores for other values of these hyperparameters in Appendix A. We used default values for other hyperparameters of GPT-3 models.

We also measured few-shot learning abilities of BioBERT on the five BioNLP tasks. We utilized both Base and Large models, each one having 110M and 340M parameters, respectively. The models were already pretrained on two large biomedical text corpora, i.e. PubMed and PMC. We experimented with different numbers of few-shot examples randomly sampled from the training set. The full list of hyperparameter values is presented in Appendix B. Table 2 presents the scores obtained by BioBERT in different few-shot settings.

As Table 1 and Table 2 demonstrate, GPT-3 and BioBERT could not obtain few-shot performance comparable to a language model fine-tuned on the whole training set in the BioNLP domain. Although few-shot performance in our experiments is far from state-of-the-art, previous work showed that, in few-shot settings, GPT-3 can achieve results competitive with state-of-the-art in open-domain NLP, even surpassing powerful fine-tuned models on some tasks (Brown et al., 2020). This suggests that such universal multi-task language models may not have enough capability to master BioNLP tasks in few-shot settings. More in-context examples, and maybe parameter modifications, might be needed for proper domain adaption.

There were some limitations that prevented us from having performance scores for all the hyperparameter values we tested on GPT-3. First, in most experiments where we used the Completion function, we could not test more than 50 few-shot learning samples per query because the length of context window could not exceed 2048 tokens. Second, when we used Classification and Answer functions, we could not get results when less than 20 few-shot learning samples (even more than 20 for some tasks) were used, because these two functions could not find enough relevant examples to predict a label or an answer.

As can be seen in Table 2, the larger BioBERT model performed better than the smaller one. Moreover, the performance generally improved by increasing the number of few-shot learning samples. These results are in line with previous findings that few-shot performance increases with model capacity and number of in-context examples (Brown et al., 2020).

| Language model | #training samples | BioNLP tasks | | | | |
|---|---|---|---|---|---|---|
| | | MedNLI | BioText | MedSTS | PubMed-RCT | PubMed-QA |
| BioBERT-Base | 5 | 0.36 | 0.49 | 0.38 | 0.51 | 0.44 |
| | 10 | 0.41 | 0.54 | 0.41 | 0.55 | 0.49 |
| | 20 | 0.45 | 0.60 | 0.49 | 0.57 | 0.56 |
| | 50 | 0.49 | 0.63 | 0.52 | 0.61 | 0.61 |
| | 100 | 0.55 | 0.68 | 0.57 | 0.69 | 0.65 |
| | 200 | 0.62 | 0.72 | 0.61 | 0.75 | 0.69 |
| BioBERT-Large | 5 | 0.39 | 0.52 | 0.41 | 0.51 | 0.45 |
| | 10 | 0.42 | 0.57 | 0.44 | 0.59 | 0.47 |
| | 20 | 0.48 | 0.61 | 0.48 | 0.63 | 0.52 |
| | 50 | 0.51 | 0.68 | 0.51 | 0.65 | 0.58 |
| | 100 | 0.63 | 0.73 | 0.60 | 0.71 | 0.66 |
| | 200 | 0.67 | 0.75 | 0.63 | 0.77 | 0.70 |
| **Fine-tuned BioBERT-Large** | Full training set | 0.85 | 0.91 | 0.84 | 0.92 | 0.81 |

Table 2: Performance of BioBERT on the BioNLP tasks in few-shot settings using Base and Large language models.

## 4 Related work

Previous NLP models mostly relied on learning task-specific representations or utilizing universal word embeddings in task-specific neural network architectures (Mikolov et al., 2013; Peters et al., 2018). Advances in task-agnostic pretraining then led to large language models providing task-agnostic representations that could be fine-tuned to enable transfer learning to specific tasks (Devlin et al., 2018; Yang et al., 2019). Recent work focused on developing multi-task language models that need few or even zero training samples, also little or even no parameter and architecture modification, to adapt to new tasks (Brown et al., 2020; Radford et al., 2019). This can be very helpful in low-resource tasks where only limited training data are available.

Only limited work on few-shot learning in the BioNLP domain has been done so far. Research mostly focused on adding changes to different steps of learning procedure to make neural networks suitable for few-shot scenarios. One previous work (Hofer et al., 2018) introduced sequential improvements on learning capabilities of a neural network in different parts of its architecture, so that few-shot performance on medical named entity recognition improved. Few-shot learning was already utilized to post-process pretrained representations, with the help of concepts contained in biomedical ontologies, in order to generate robust biomedical name representations (Fivez et al., 2021).

## 5 Conclusion

In this paper, we investigated few-shot transfer learning abilities of two large language models, i.e. GPT-3 and BioBERT, on several BioNLP tasks. As the experimental results demonstrated, these language models are not proficient few-shot learners in the biomedical domain. Future work should target extensive pretraining of GPT-3 on biomedical text, as we saw that BioBERT-Large performed better in few-shot settings, although it had 514 times fewer parameters than GPT-3. However, large-scale pretraining of GPT-3 requires extreme amounts of computational power and consumes a lot of electricity (Brown et al., 2020); not every research group has access to highly powerful computing resources.

Given that BioBERT could not perform well in our few-shot experiments, pretraining on biomedical texts does not seem to be sufficient for achieving sufficient few-shot performance. Modifications to the transformer architecture, as well as to pretraining and few-shot learning methodologies seem to be required. Task-specific architectures and training strategies may also provide effective solutions to in-domain, task-specific few-shot knowledge transfer.

In those tasks that required comparing two sentences, i.e. MedNLI and MedSTS, GPT-3 performed worse than in other tasks. Similar findings were already reported on open-domain texts (Brown et al., 2020). Moreover, GPT-3 does not benefit from bidirectional architecture. Further methodological improvements would be necessary to enhance GPT-3's few-shot performance.

**Appendix A. GPT-3 performance scores for other values of *max_examples* and *max_rerank***

The following tables present GPT-3 performnace scores for other values of hyperparameters *max_examples* and *max_rerank*.

*max_examples* controlled the maximum number of relevant examples extracted by GPT-3 from few-shot learning samples to predict most probable labels when the `Classification` function was used.

*max_rerank* controlled the maximum number of relevant documents extracted by GPT-3 from few-shot learning samples to predict most probable answers when the `Answer` function was used.

| GPT-3 built-in function | #training samples | BioNLP tasks | | | | |
|---|---|---|---|---|---|---|
| | | MedNLI | BioText | MedSTS | PubMed-RCT | PubMed-QA |
| Classification | 5 | ** | ** | ** | ** | - |
| | 10 | ** | ** | ** | ** | - |
| | 20 | 0.28 | ** | ** | 0.29 | - |
| | 50 | 0.31 | 0.35 | 0.16 | 0.37 | - |
| | 100 | 0.37 | 0.49 | 0.13 | 0.41 | - |
| | 200 | 0.35 | 0.51 | 0.27 | 0.40 | - |
| Answer | 5 | - | - | - | - | ** |
| | 10 | - | - | - | - | ** |
| | 20 | - | - | - | - | 0.43 |
| | 50 | - | - | - | - | 0.51 |
| | 100 | - | - | - | - | 0.57 |
| | 200 | - | - | - | - | 0.52 |

Table 3: Performance of GPT-3 on the BioNLP tasks in few-shot settings using two different built-in functions. These scores were obtained for *max_examples*=2 and *max_rerank*=2. *The length of prompt exceeds the model's context window that equals 2048 tokens. **The model could not find enough relevant training samples for several test samples, therefore, no scores were produced because of a model's error.

| GPT-3 built-in function | #training samples | BioNLP tasks | | | | |
|---|---|---|---|---|---|---|
| | | MedNLI | BioText | MedSTS | PubMed-RCT | PubMed-QA |
| Classification | 5 | ** | ** | ** | ** | - |
| | 10 | ** | ** | ** | ** | - |
| | 20 | 0.29 | ** | ** | 0.30 | - |
| | 50 | 0.31 | 0.36 | 0.17 | 0.37 | - |
| | 100 | 0.38 | 0.48 | 0.12 | 0.42 | - |
| | 200 | 0.34 | 0.51 | 0.25 | 0.40 | - |
| Answer | 5 | - | - | - | - | ** |
| | 10 | - | - | - | - | ** |
| | 20 | - | - | - | - | 0.41 |
| | 50 | - | - | - | - | 0.52 |
| | 100 | - | - | - | - | 0.56 |
| | 200 | - | - | - | - | 0.50 |

Table 4: Performance of GPT-3 on the BioNLP tasks in few-shot settings using two different built-in functions. These scores were obtained for *max_examples*=3 and *max_rerank*=3. *The length of prompt exceeds the model's context window that equals 2048 tokens. **The model could not find enough relevant training samples for several test samples, therefore, no scores were produced because of a model's error.

| GPT-3 built-in function | #training samples | BioNLP tasks | | | | |
|---|---|---|---|---|---|---|
| | | MedNLI | BioText | MedSTS | PubMed-RCT | PubMed-QA |
| Classification | 5 | ** | ** | ** | ** | - |
| | 10 | ** | ** | ** | ** | - |
| | 20 | 0.28 | ** | ** | ** | - |
| | 50 | 0.33 | ** | 0.18 | 0.35 | - |
| | 100 | 0.39 | 0.47 | 0.14 | 0.41 | - |
| | 200 | 0.35 | 0.52 | 0.26 | 0.42 | - |
| Answer | 5 | - | - | - | - | ** |
| | 10 | - | - | - | - | ** |
| | 20 | - | - | - | - | 0.41 |
| | 50 | - | - | - | - | 0.53 |
| | 100 | - | - | - | - | 0.57 |
| | 200 | - | - | - | - | 0.52 |

Table 5: Performance of GPT-3 on the BioNLP tasks in few-shot settings using two different built-in functions. These scores were obtained for *max_examples*=4 and *max_rerank*=4. *The length of prompt exceeds the model's context window that equals 2048 tokens. **The model could not find enough relevant training samples for several test samples, therefore, no scores were produced because of a model's error.

| GPT-3 built-in function | #training samples | BioNLP tasks | | | | |
|---|---|---|---|---|---|---|
| | | MedNLI | BioText | MedSTS | PubMed-RCT | PubMed-QA |
| Classification | 5 | ** | ** | ** | ** | - |
| | 10 | ** | ** | ** | ** | - |
| | 20 | 0.31 | ** | ** | ** | - |
| | 50 | 0.33 | ** | ** | ** | - |
| | 100 | 0.41 | ** | 0.14 | 0.45 | - |
| | 200 | 0.38 | 0.54 | 0.28 | 0.42 | - |
| Answer | 5 | - | - | - | - | ** |
| | 10 | - | - | - | - | ** |
| | 20 | - | - | - | - | 0.43 |
| | 50 | - | - | - | - | 0.54 |
| | 100 | - | - | - | - | 0.60 |
| | 200 | - | - | - | - | 0.52 |

Table 6: Performance of GPT-3 on the BioNLP tasks in few-shot settings using two different built-in functions. These scores were obtained for *max_examples*=6 and *max_rerank*=6. *The length of prompt exceeds the model's context window that equals 2048 tokens. **The model could not find enough relevant training samples for several test samples, therefore, no scores were produced because of a model's error.

| GPT-3 built-in function | #training samples | BioNLP tasks | | | | |
|---|---|---|---|---|---|---|
| | | MedNLI | BioText | MedSTS | PubMed-RCT | PubMed-QA |
| Classification | 5 | ** | ** | ** | ** | - |
| | 10 | ** | ** | ** | ** | - |
| | 20 | 0.32 | ** | ** | ** | - |
| | 50 | 0.32 | ** | ** | ** | - |
| | 100 | 0.41 | ** | 0.11 | 0.44 | - |
| | 200 | 0.38 | 0.52 | 0.25 | 0.43 | - |
| Answer | 5 | - | - | - | - | ** |
| | 10 | - | - | - | - | ** |
| | 20 | - | - | - | - | 0.44 |
| | 50 | - | - | - | - | 0.55 |
| | 100 | - | - | - | - | 0.59 |
| | 200 | - | - | - | - | 0.53 |

Table 7: Performance of GPT-3 on the BioNLP tasks in few-shot settings using two different built-in functions. These scores were obtained for *max_examples*=7 and *max_rerank*=7. *The length of prompt exceeds the model's context window that equals 2048 tokens. **The model could not find enough relevant training samples for several test samples, therefore, no scores were produced because of a model's error.

| GPT-3 built-in function | #training samples | BioNLP tasks ||||| 
| | | MedNLI | BioText | MedSTS | PubMed-RCT | PubMed-QA |
|---|---|---|---|---|---|---|
| Classification | 5 | ** | ** | ** | ** | - |
| | 10 | ** | ** | ** | ** | - |
| | 20 | 0.31 | ** | ** | ** | - |
| | 50 | 0.33 | ** | ** | ** | - |
| | 100 | 0.41 | ** | 0.13 | 0.45 | - |
| | 200 | 0.38 | 0.53 | 0.27 | 0.42 | - |
| Answer | 5 | - | - | - | - | ** |
| | 10 | - | - | - | - | ** |
| | 20 | - | - | - | - | 0.41 |
| | 50 | - | - | - | - | 0.55 |
| | 100 | - | - | - | - | 0.60 |
| | 200 | - | - | - | - | 0.52 |

Table 8: Performance of GPT-3 on the BioNLP tasks in few-shot settings using two different built-in functions. These scores were obtained for *max_examples*=8 and *max_rerank*=8. *The length of prompt exceeds the model's context window that equals 2048 tokens. **The model could not find enough relevant training samples for several test samples, therefore, no scores were produced because of a model's error.

| GPT-3 built-in function | #training samples | BioNLP tasks |||||
| | | MedNLI | BioText | MedSTS | PubMed-RCT | PubMed-QA |
|---|---|---|---|---|---|---|
| Classification | 5 | ** | ** | ** | ** | - |
| | 10 | ** | ** | ** | ** | - |
| | 20 | 0.29 | ** | ** | ** | - |
| | 50 | 0.34 | ** | ** | ** | - |
| | 100 | 0.41 | ** | 0.12 | 0.44 | - |
| | 200 | 0.39 | 0.54 | 0.28 | 0.41 | - |
| Answer | 5 | - | - | - | - | ** |
| | 10 | - | - | - | - | ** |
| | 20 | - | - | - | - | 0.40 |
| | 50 | - | - | - | - | 0.56 |
| | 100 | - | - | - | - | 0.58 |
| | 200 | - | - | - | - | 0.53 |

Table 9: Performance of GPT-3 on the BioNLP tasks in few-shot settings using two different built-in functions. These scores were obtained for *max_examples*=9 and *max_rerank*=9. *The length of prompt exceeds the model's context window that equals 2048 tokens. **The model could not find enough relevant training samples for several test samples, therefore, no scores were produced because of a model's error.

| GPT-3 built-in function | #training samples | BioNLP tasks |||||
| | | MedNLI | BioText | MedSTS | PubMed-RCT | PubMed-QA |
|---|---|---|---|---|---|---|
| Classification | 5 | ** | ** | ** | ** | - |
| | 10 | ** | ** | ** | ** | - |
| | 20 | 0.31 | ** | ** | ** | - |
| | 50 | 0.33 | ** | ** | ** | - |
| | 100 | 0.39 | ** | 0.12 | 0.43 | - |
| | 200 | 0.38 | 0.54 | 0.27 | 0.40 | - |
| Answer | 5 | - | - | - | - | ** |
| | 10 | - | - | - | - | ** |
| | 20 | - | - | - | - | 0.43 |
| | 50 | - | - | - | - | 0.55 |
| | 100 | - | - | - | - | 0.57 |
| | 200 | - | - | - | - | 0.50 |

Table 10: Performance of GPT-3 on the BioNLP tasks in few-shot settings using two different built-in functions. These scores were obtained for *max_examples*=10 and *max_rerank*=10. *The length of prompt exceeds the model's context window that equals 2048 tokens. **The model could not find enough relevant training samples for several test samples, therefore, no scores were produced because of a model's error.

# Appendix B. Hyperparameter values of BioBERT

Table 11 presents the hyperparameter values of the model BioBERT-`Large` that was fine-tuned on the whole training set. Table 12 presents the hyperparameter values of the model BioBERT-`Large` utilized in the few-shot experiments. Table 13 presents the hyperparameter values of the model BioBERT-`Base` utilized in the few-shot experiments.

| Hyperparameter | PubMed-RCT | MedNLI | PubMed-QA | BioText | Med-STS |
| --- | --- | --- | --- | --- | --- |
| Max sequence length | 64 | 64 | 256 | 64 | 64 |
| Batch size | 12 | 12 | 2 | 12 | 12 |
| Learning rate | 1e-5 | 1e-5 | 1.5e-5 | 2e-5 | 2e-5 |
| Weight decay | 0.1 | 0.1 | 0.01 | 0.1 | 0.01 |
| Learning rate decay | Linear | Linear | Linear | Linear | Linear |
| Warmup ratio | 0.06 | 0.06 | 0.06 | 0.05 | 0.05 |
| Num epochs | 20 | 15 | 10 | 20 | 20 |

Table 11: Hyperparameter values of the model BioBERT-`Large` fine-tuned on the whole training set.

| Hyperparameter | PubMed-RCT | MedNLI | PubMed-QA | BioText | Med-STS |
| --- | --- | --- | --- | --- | --- |
| Max sequence length | 64 | 64 | 256 | 64 | 64 |
| Batch size | 12 | 12 | 2 | 12 | 12 |
| Learning rate | 2e-5 | 2e-5 | 2.5e-5 | 3e-5 | 3e-5 |
| Weight decay | 0.01 | 0.01 | 0.01 | 0.01 | 0.001 |
| Learning rate decay | Linear | Linear | Linear | Linear | Linear |
| Warmup ratio | 0.08 | 0.08 | 0.08 | 0.08 | 0.08 |
| Num epochs | 5 | 5 | 5 | 5 | 5 |

Table 12: Hyperparameter values of the model BioBERT-`Large` utilized in the few-shot experiments.

| Hyperparameter | PubMed-RCT | MedNLI | PubMed-QA | BioText | Med-STS |
| --- | --- | --- | --- | --- | --- |
| Max sequence length | 64 | 64 | 256 | 64 | 64 |
| Batch size | 12 | 12 | 2 | 12 | 12 |
| Learning rate | 3e-5 | 3e-5 | 3e-5 | 3e-5 | 3e-5 |
| Weight decay | 0.01 | 0.01 | 0.01 | 0.01 | 0.01 |
| Learning rate decay | Linear | Linear | Linear | Linear | Linear |
| Warmup ratio | 0.08 | 0.08 | 0.08 | 0.08 | 0.08 |
| Num epochs | 5 | 5 | 5 | 5 | 5 |

Table 13: Hyperparameter values of the model BioBERT-`Base` utilized in the few-shot experiments.